\pdfoutput=1

\documentclass[11pt]{article}
\usepackage{acl}

\usepackage{tikz}
\usepackage{pgfplots}
\usepackage{pgfplotstable}
\usepackage{pgf-pie}
\usepackage{subcaption}

\usepackage{xcolor}
\definecolor{model4o}{RGB}{31,119,180}       
\definecolor{modelgem25}{RGB}{231,21,255}    
\definecolor{modelgem2}{RGB}{152,0,168}      
\definecolor{modelr1}{RGB}{214,39,40}        
\definecolor{modelson35}{RGB}{255,138,21}   
\definecolor{modelson37}{RGB}{255,178,102}     
\definecolor{modelv3}{RGB}{227,119,194}      

\usepackage{latexsym}
\usepackage{booktabs}
\usepackage{adjustbox}
\usepackage{multirow}
\pgfplotsset{compat=1.18}
\usepackage{amssymb} 
\usepackage{pifont}  
\usepackage{longtable}
\usepackage{array}
\usepackage{adjustbox}

\usepackage{fvextra}            
\usepackage{listings}           
\usepackage[most]{tcolorbox}
\tcbuselibrary{skins,breakable,listings}

\newtcolorbox{PromptBox}[2][]{%
  enhanced,
  breakable,
  colback=gray!3,
  colframe=black!20,
  title={#2},
  fonttitle=\bfseries,
  listing only,
  listing engine=listings,
  listing options={
    basicstyle=\ttfamily\small,
    breaklines=true,
    breakatwhitespace=false,
    columns=fullflexible,
    upquote=true,
    showstringspaces=false,
    tabsize=2
  },
  #1
}

\newcommand{\cmark}{\ding{51}}  
\newcommand{\xmark}{\ding{55}}  

\usepackage{inconsolata}
\usepackage{graphicx}

\title{Beyond One World: Benchmarking Super Heros in Role-Playing Across Multiversal Contexts}

\author{
\textbf{Perapard Ngokpol\textsuperscript{1}\thanks{Equal contribution.}},
\textbf{Kun Kerdthaisong\textsuperscript{1}\footnotemark[1]},
\textbf{Pasin Buakhaw\textsuperscript{2}\footnotemark[1]},\textbf{Pitikorn Khlaisamniang\textsuperscript{3}},\\
\textbf{Supasate Vorathammathorn\textsuperscript{3}},\textbf{Piyalitt Ittichaiwong\textsuperscript{4,5}\thanks{Corresponding authors.}},
\textbf{Nutchanon Yongsatianchot\textsuperscript{1}\footnotemark[2]}\\[4pt]
\textsuperscript{1}Faculty of Engineering, Thammasat School of Engineering, Thammasat University\\
\textsuperscript{2}Department of Computer Engineering and Digital Technology, Faculty of Engineering, Chulalongkorn University\\
\textsuperscript{3}Artificial Intelligence Association of Thailand\\
\textsuperscript{4}School of Biomedical Engineering \& Imaging Sciences, King’s College London\\
\textsuperscript{5}Siriraj Informatics and Data Innovation Center (SIData+), \\Faculty of Medicine, Siriraj Hospital, Mahidol University
}

\begin{document}
\maketitle

\begin{abstract}
Large language models (LLMs) are increasingly used as role-playing agents, yet their capacity to faithfully and consistently portray version-specific characters—for example, superheroes across comic and cinematic universes—remains underexplored. Superhero canons such as Marvel and DC provide a rich testbed: decades of storytelling yield multiple incarnations of the same character with distinct histories, values, and moral codes. To study this problem, we introduce \textit{Beyond One World}, a benchmark for character-grounded role-play spanning 30 iconic heroes and 90 canon-specific versions. The benchmark comprises two tasks: \textbf{(i)} Canon Events, which probes factual recall of pivotal life stages, and \textbf{(ii)} Moral Dilemmas, which confronts models with ethically charged scenarios. We score responses for canonical accuracy and reasoning fidelity under a framework that separates internal deliberation (``thinking'') from outward decisions (``acting''). We further propose \textbf{Think--Act Matching}, a metric that quantifies alignment between reasons and actions and serves as a proxy for model trustworthiness. Experiments across reasoning- and non-reasoning-oriented models yield three findings: \textbf{(1)} chain-of-thought prompting improves narrative coherence in weaker models but can reduce canonical accuracy in stronger ones; \textbf{(2)} cross-version generalization within a character remains a major obstacle; and \textbf{(3)} models often excel at either thinking or acting, but rarely both. \textit{Beyond One World} exposes critical gaps in multiversal consistency and reasoning alignment, offering a challenging evaluation for role-playing LLMs.

\footnotetext[1]{The GitHub project is available at \url{https://github.com/Augustus2011/Beyond_One_World}.}

\footnotetext[2]{Datasets are available at \url{https://huggingface.co/collections/Character-lab/emlp-wordplay-2025-68dcdbff705d8004c5d03087}}


\end{abstract}

\section{Introduction}
\label{1.Introduction}

Large language models (LLMs) have demonstrated strong performance in text generation, translation, and increasingly complex reasoning tasks~\citep{lu-etal-2024-large}. Their growing proficiency has enabled sophisticated applications, including the simulation of personalities and characters~\citep{xi2023risepotentiallargelanguage}. In \emph{character-based role-playing} (C\mbox{-}RP), models adopt given personas and are expected to emulate the target's knowledge, speaking style, and behavior~\citep{chen2025designguidelinerpaevaluation}.

Recent advances in explicit reasoning---such as chain-of-thought (CoT) prompting~\citep{COT} and specialized reasoning models like DeepSeek-R1~\citep{r1} and Gemini 2.5 Thinking~\citep{google2025gemini2.5}---further extend LLM capabilities. However, the application of these methods to nuanced role-play remains underexplored~\citep{reasoningdoesnecessarilyimprove}. A central challenge is whether LLMs can \emph{consistently portray version-specific characters} across contexts~\citep{wang-etal-2024-incharacter}. Existing benchmarks typically target single-character consistency or basic factual recall. While models can convincingly inhabit a persona in isolated settings, it is unclear whether they maintain coherence as that persona evolves across a narrative timeline---for example, portraying a younger, idealistic incarnation versus a later, more cynical one~\citep{ahn-etal-2024-timechara}. Moreover, current evaluations rarely analyze the alignment between a model's \emph{reasoning traces} and its \emph{decisions}, across both reasoning-oriented and standard models. Notably, even standard models often produce implicit reasoning (e.g., inner monologue), motivating a more fine-grained assessment of cognitive cues in generated text.

To address these gaps, we focus on superhero characters from the \emph{Marvel} and \emph{DC} universes, which offer well-documented, canonically distinct identities spanning timelines and parallel worlds~\citep{stammbach-etal-2022-heroes}. Decades of storytelling provide detailed histories, evolving moral codes, psychological profiles, motivations, and relationships for numerous versions of the same hero. This rich, curated narrative record yields a complex yet comparatively stable ground truth for evaluating multiversal, version-specific role-play.

In this work, we introduce \textit{Beyond One World}, a benchmark for character-grounded role-play covering 30 iconic heroes and 90 canon-specific versions. The benchmark assesses two complementary dimensions: \textbf{(i)} \emph{Canon Events}, which probes factual recall at pivotal life stages (childhood, pre-hero, and established-hero phases) via multiple-choice questions; and \textbf{(ii)} \emph{Moral Dilemmas}, which presents ethically charged situations inspired by established narrative themes (e.g., \emph{save one vs.\ the greater good}, \emph{duty vs.\ personal desire}, \emph{moral code vs.\ ends-justify-the-means}). We evaluate responses with an LLM-as-a-judge rubric that separates internal deliberation (``thinking'') from outward decisions (``acting'') and score both canonical accuracy and reasoning fidelity.

Our contributions are as follows: 
\begin{itemize}
    \item \textbf{Dataset Benchmark:} We release \textit{Beyond One World}, a dataset for version-specific character role-play across 30 heroes and 90 versions, with two tasks—\emph{Canon Events} and \emph{Moral Dilemmas}—targeting factual recall and ethically grounded decision-making.
    \item \textbf{Evaluation framework:} We propose an analysis framework that disentangles ``thinking'' from ``acting'' and applies to both reasoning-oriented and standard LLMs, enabling fine-grained assessment of character-consistent reasoning and decisions.
\end{itemize}

\begin{table*}[ht]
\centering
\label{tab:dataset_comparison}
\resizebox{\textwidth}{!}{%
\begin{tabular}{lccccccp{5cm}}
\toprule
\textbf{Dataset/Benchmark} &
\textbf{Auto-Generated} &
\textbf{Point-in-time} &
\textbf{Multiversal Context} &
\textbf{Reasoning Analysis} &
\textbf{Items} \\
\textbf{} &
\textbf{} &
\textbf{role-playing} &
\textbf{same character(movie,series,novel)} &
\textbf{} &
\textbf{} &
\textbf{} \\
\midrule
SocialBench~\cite{socialbench} & \cmark & \xmark & \xmark & \xmark & 6,493 \\
RoleBench~\cite{wang-etal-2024-rolellm} & \cmark & \xmark & \xmark & \xmark & 34,523 \\
HPD~\cite{HPD_dataset} & \xmark & \cmark & \xmark & \xmark & 316 \\
CharacterEval~\cite{charactereval} & \cmark & \xmark & \xmark &\xmark & 4,564 \\
TimeChara~\cite{ahn-etal-2024-timechara} & \cmark & \cmark & \xmark & \xmark &  10,895 \\
\textbf{Beyond One World(Ours)} & \cmark & \xmark & \cmark &\cmark & 2,426 \\
\bottomrule
\end{tabular}
}
\caption{Comparison of Benchmarks}
\end{table*}

\section{Related Work}
\label{2.Related Work}
Large language models (LLMs) have recently been used as \textbf{role-playing language agents} (RPLAs) that converse while inhabiting coherent personas. Below, we summarize the main strands of prior work that inform our benchmark.

\subsection{From Persona-Chat to Modern RPLAs}
Early persona-grounded dialogue framed the task as conditioning a chatbot on a short textual biography. The \textsc{PersonaChat} corpus showed that even simple profiles improve response consistency over generic chit-chat models~\citep{zhang-etal-2018-personalizing}. Subsequent work explored richer conditioning signals—such as goals, memories, and affect—laying the groundwork for contemporary RPLAs.

\subsection{Character-Alignment Corpora}
More recent datasets target \emph{character} personas with denser annotations. The \textbf{Harry Potter Dialogue} (HPD) corpus pairs approximately 1k multi-turn scenes with scene metadata, speaker roles, and evolving inter-character relations in English and Chinese, enabling fine-grained character imitation~\citep{HPD_dataset}. In Chinese, \textbf{CharacterEval} provides 1{,}785 dialogues for 77 literary figures and introduces thirteen metrics spanning conversational ability, consistency, attractiveness, and personality back-testing~\citep{charactereval}.

\subsection{Evaluating Sociality and Personalization}
Beyond single-speaker fidelity, \textbf{SocialBench} benchmarks 500 personas across roughly 6k scenarios, probing self-awareness, emotion perception, and group adaptability; strong solo performance does not necessarily translate to robust group behavior~\citep{socialbench}. Complementing this, \textbf{PersoBench} measures how well LLMs tailor responses to user-supplied profiles across three dialogue corpora, revealing substantial gaps in personalization despite fluent generation~\citep{PersoBench}.

\subsection{Temporal Consistency on Characters}
Maintaining a persona’s limited knowledge at a specific point in time remains challenging. \textbf{TimeChara} evaluates ``point-in-time'' hallucinations across 10{,}895 instances and shows that even GPT-4o can leak future facts or inconsistent traits; narrative-expert decomposition mitigates but does not eliminate the issue~\citep{ahn-etal-2024-timechara}. Our benchmark extends this line by testing multiple \emph{universe variants} (e.g., childhood, adolescence, pre-hero, hero) of the same character, stressing both temporal and cross-timeline coherence.

\section{Dataset Creation}
\label{3.Dataset Creation}

We manually curated a set of well-known \emph{hero} characters from the Marvel and DC universes, drawing on films, comics, and television series. To avoid role ambiguity, we excluded arcs in which a chosen figure adopts a villainous identity see more details in Appendix~\ref{Question Constructions}. The final collection comprises \textbf{30 distinct heroes}, each represented in three narrative variants (childhood, pre-hero, and hero phase), yielding \textbf{90 character versions} in total.

To evaluate an LLM’s factual recall and moral reasoning, we organized the corpus into two tasks:  
\textbf{(i)} \textbf{Canonical Events} (\ref{Canon Events}) probes knowledge of pivotal moments in each hero’s timeline;  
\textbf{(ii)} \textbf{Moral Dilemma Situations} (\ref{Dilemma Situations}) challenge the model to choose actions consistent with the hero’s ethical code;

\subsection{Canon Events}
\label{Canon Events}

This task tests whether an LLM can faithfully recall the key events that define a hero’s backstory.  Human annotators with domain expertise wrote four–option multiple-choice questions anchored to pivotal moments in each character’s lore.  To reflect a hero’s narrative arc, we split the timeline into three phases: \textbf{Childhood}, spanning early life through adolescence; \textbf{Pre-Hero}, covering the period in which powers are acquired or the first steps toward heroism are taken; and the active \textbf{Hero phase}. For every timeline variant we supply three questions on Childhood, three on Pre-Hero, and nine on the Hero phase, yielding a total of \(1{,}346\) hand-curated items.

\subsection{Dilemma Situations}
\label{Dilemma Situations}

To gauge whether an LLM can deliberate in line with a hero's ethical compass, we created a bank of \textbf{1,080 multiple-choice dilemmas} through a systematic generation and curation process. For each of the 90 character versions, we synthesized three distinct scenarios spanning four archetypal moral conflicts using \textbf{Claude Sonnet 3.7}, then manually filtered them for canonical fidelity and linguistic clarity.

\subsubsection{Dilemma Generation Methodology}
Our dilemma generation process employed a structured prompt-based approach designed to ensure both character-specific authenticity and scenario diversity. The generation pipeline consisted of several key components:

\textbf{Character Integration:} Each dilemma was grounded in the specific character's background, abilities, and established moral framework as derived from their canonical source material. The generation prompt explicitly incorporated the character's name, lore source, and contextual background to ensure scenario relevance and authenticity.

\textbf{Iterative Diversification:} To prevent redundancy and ensure rich scenario variation, we implemented an iterative generation strategy. For each character-dilemma type combination, three distinct scenarios were created sequentially, with each subsequent iteration explicitly instructed to avoid similarities to previously generated scenarios for that character. This approach yielded scenarios that maintained thematic consistency while exploring different manifestations of the core moral conflict.

\textbf{Output:} Each generated dilemma included:(1) a situational description establishing context and stakes, (2) two binary choice options representing competing moral imperatives, and (3) explicit consequences for each choice. See prompt in (Appendix~\ref{Dilemma Generation})

\subsubsection{Moral Conflict Taxonomy}

We selected four archetypal conflicts that commonly occur in superhero narratives and capture fundamental tensions in moral reasoning across diverse ethical frameworks:

\textbf{1. Save One vs. Save the Greater Good} probes a character's willingness to sacrifice a few for the many; Iron Man's self-sacrifice in \textit{Avengers: Endgame} exemplifies this tension and echoes existential meaning-making through sacrifice~\citep{frankl2006man} as well as Kohlberg's post-conventional stage of moral reasoning~\citep{kohlberg1971stages}. Generated scenarios in this category typically involved time-sensitive situations where characters must choose between saving a beloved individual versus protecting a larger population, forcing a direct confrontation between particularistic loyalties and universalistic obligations.

\textbf{2. Hero or Villain} captures the struggle between a virtuous self-image and darker impulses—Anakin Skywalker's fall from grace being the archetype—and draws on Jung's shadow theory~\citep{jung1959archetypes}, Bandura's moral disengagement~\citep{bandura1999moral}, and Eriksonian identity development~\citep{erikson1968identity}. These scenarios explored moments of moral temptation where characters face opportunities to achieve desired outcomes through morally questionable means, testing their commitment to heroic ideals under pressure.

\textbf{3. Duty vs. Personal Desire} mirrors Spider-Man's dual life and invokes cognitive dissonance~\citep{festinger1957cognitive}, recurring psychosocial role conflict~\citep{erikson1968identity}, and Maslow's hierarchy of needs~\citep{maslow1943theory}. Generated dilemmas in this category presented characters with situations where their heroic responsibilities directly conflicted with personal relationships, aspirations, or well-being, examining how they prioritize competing life domains.

\textbf{4. Ends Justify the Means vs. Moral Code} tests whether utilitarian outcomes override deontological constraints; Batman's refusal to kill despite expedient benefits illustrates the clash between Kantian ethics~\citep{kant1785groundwork} and the risk of moral injury or ethical fatigue~\citep{litz2009moral}. These scenarios forced characters to choose between adhering to their established moral principles or compromising them to achieve objectively beneficial outcomes.

\begin{table}[ht]
\centering

\begin{tabular}{ll}
\toprule
\textbf{Task} & \textbf{Datapoints} \\
\midrule
\textbf{Canon} & (270, 270, 806) \\
\textbf{Dilemma} & (270, 270, 270, 270) \\
\bottomrule
\end{tabular}
\label{tab:dataset_statistics}
\caption{summary of our benchmark \textbf{Task 1} consist of 270 Childhood, 270 Pre-Hero and 806 Hero questions(e.g.,Hero1, Hero2, Hero3) while \textbf{Task 2} consist of balanced Dilemma situations.}
\end{table}

\begin{figure}
    \centering
    \includegraphics[width=1.0\linewidth]{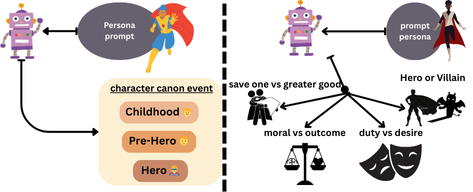}
    \caption{Left image is inferencing llm that prompted hero persona to do canon event task, and the right image is prompted llm to do dilemma situation task.}
    \label{fig:HEROCharacter_INFO}
\end{figure}

\section{Experimental Setups}
\label{Experimental Setups}

\subsection{Main Experiment}
\label{Main Experiment}
After dataset creation, we designed experiments to evaluate how effectively large language models can adopt character role-playing on hero characters As (Figure~\ref{fig:HEROCharacter_INFO}). Our experimental framework mainly focus on \textbf{multiple choice question answering} with the structured prompting strategies as bellow. 

\subsubsection{Structured Prompting Design}
\label{Structured Prompting Design}
We developed task-specific prompt templates that provide minimal but essential character context to the models. We employ a consistent base structure.
\begin{PromptBox}{Initial hero persona}
    You are playing the role of \textbf{$<$name$>$}, act and think as \textbf{$<$name$>$}, from \textbf{$<$lore$>$} $<$question$>$ \textbf{{formatted\_question}} $<$question/$>$
\end{PromptBox}

The $<$name$>$ is character name such \textit{Stark “Iron Man”} while $<$lore$>$ is source material as the primary identity anchors. lastly formatted\_question depends on which task. See Full Task prompt in (Appendix~\ref{Task prompt}).

\subsubsection{Cross-Character Evaluation}
\label{Cross-Character Evaluation}
To assess the robustness of character differentiation, we implemented a cross-character evaluation protocol. This approach leverages characters that share the same name but originate from different fictional universes (e.g., different movie adaptations, comic series, or TV shows of the same character).
\paragraph{Example} Given a set of characters with identical names but distinct source materials ($CID_4$, $CID_5$, $CID_6$), we systematically evaluate cross-character.
\begin{PromptBox}{Cross-Character Evaluation}
Character Set: Spider-Man variants\\
- $CID_4$: Spider-Man (Marvel Cinematic Universe, 2016)\\
- $CID_5$: Spider-Man (The Amazing Spider-Man 2, 2014)\\
- $CID_6$: Spider-Man (Sam Raimi Films, 2002)\\
Evaluation: $CID_4$ answers questions originally designed for $CID_5$ and $CID_6$    
\end{PromptBox}

\subsection{Reasoning analysis framework}
\label{Reasoning analysis framework}
We introduce a reasoning analysis framework. Given task outputs from Canon (Section~\ref{Canon Events}) and Dilemma (Section~\ref{Dilemma Situations}), responses are first segmented into two reasoning traits: \texttt{$<$thinking$>$} capturing internal deliberation, and \texttt{$<$acting$>$} capturing decision-making behavior or Physical Appearance. This structured output by Gpt-4o-mini is then judged by Sonnet 3.7, which is prompted to act as a judge scoring. 

\begin{figure}[ht]
    \centering
    \includegraphics[width=1.0\linewidth]{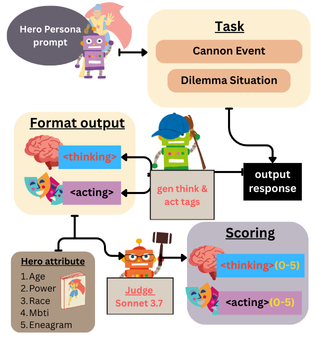}
    \caption{A pipeline for reasoning analysis, after getting output from task Canon and Dilemma the output response are structured into $<$thinking$>$ and $<$acting$>$ by Gpt4o-mini judged by Sonnet3.7 that was prompted to be a judge with attributes of that character.}
    \label{fig:Judge_framework}
\end{figure}

To scoring the response, prompt conditions the model with character attributes (e.g., Age, Power, Race, Mbti and Enneagram) to simulate a perspective-aligned evaluation process (Appendix~\ref{Reasoning Analysis}). The score value is between 0-5. See the pipeline in (Figure \ref{fig:Judge_framework})

In addition, to quantify the alignment between a character’s internal reasoning and their outward behavior, we introduce a \textit{Think-Act Matching} procedure. We embed both \texttt{$<$thinking$>$} and \texttt{$<$acting$>$} spans with the \textbf{all-mpnet-base-v2}~\cite{all-mpnet-base-v2}. Cosine similarity is then computed between the embeddings of each pair. The maximum similarity score across pairs is selected as an indicator of how well a character’s internal reasoning aligns with its external action. This score is interpreted as a proxy for \textbf{trustworthiness} in the character’s response. See result in (Table~\ref{tab:model_comparison}).

\section{Results and Discussion}
\label{Results and Discussion}

\subsection{Overall Performance on Canon Events}
(Table~\ref{tab:model_comparison}) presents results comparing chain-of-thought (CoT) prompting with direct answering (Non-CoT) for the Canon Event task. Several patterns emerge. First, CoT prompting does not consistently improve accuracy: while models such as 4o-mini show a small positive gain (+0.02 accuracy, +0.024 F1), stronger models like sonnet3.5 and sonnet3.7 actually decline in performance when asked to verbalize intermediate reasoning. This suggests that explicit reasoning steps may introduce hallucinations or off-canon elaborations that reduce factual consistency. Interestingly, models such as r1 benefit in terms of reasoning–action alignment (cosine similarity +0.075), even as raw accuracy falls, indicating that CoT can produce more coherent internal deliberation even if the final choice is incorrect.
\begin{table*}[ht]
\small
\centering
\begin{tabular}{|l|c|c|c|c|c|c|c|c|c|}
\hline
\textbf{Model} & \multicolumn{3}{c|}{\textbf{Non-CoT}} & \multicolumn{3}{c|}{\textbf{Chain of Thought}} & \multicolumn{3}{c|}{\textbf{Difference (CoT - Non-CoT)}} \\
\cline{2-10}
 & \textbf{Acc} & \textbf{F1} & \textbf{Cosim} & \textbf{Acc} & \textbf{F1} & \textbf{Cosim} & \textbf{Acc} & \textbf{F1} & \textbf{Cosim} \\
\hline
4o-mini & 0.626 & 0.632 & \textbf{0.458} & \underline{0.646} & \underline{0.656} & \textbf{0.432} & \textcolor{green}{+0.020} & \textcolor{green}{+0.024} & \textcolor{red}{-0.026} \\
\hline
gemini2.5-flash-think & \underline{0.663} & \underline{0.665} & \underline{0.450} & 0.388 & 0.344 & 0.389 & \textcolor{red}{-0.275} & \textcolor{red}{-0.321} & \textcolor{red}{-0.061} \\
\hline
gemini2-flash & 0.638 & 0.647 & \underline{0.450} & 0.637 & 0.652 & \underline{0.408} & \textcolor{red}{-0.001} & \textcolor{green}{+0.005} & \textcolor{red}{-0.042} \\
\hline
r1 & 0.652 & 0.656 & 0.296 & 0.546 & 0.573 & 0.371 & \textcolor{red}{-0.106} & \textcolor{red}{-0.083} & \textcolor{green}{+0.075} \\
\hline
sonnet3.5 & \textbf{0.704} & \textbf{0.707} & 0.433 & \textbf{0.651} & \textbf{0.659} & 0.404 & \textcolor{red}{-0.053} & \textcolor{red}{-0.048} & \textcolor{red}{-0.029} \\
\hline
sonnet3.7 & 0.637 & 0.645 & 0.378 & 0.620 & 0.628 & 0.380 & \textcolor{red}{-0.017} & \textcolor{red}{-0.017} & \textcolor{green}{+0.002} \\
\hline
v3 & 0.615 & 0.623 & 0.423 & 0.502 & 0.497 & 0.376 & \textcolor{red}{-0.113} & \textcolor{red}{-0.126} & \textcolor{red}{-0.047} \\
\hline
\end{tabular}

\caption{Model Performance Comparison: Chain of Thought vs Non-CoT in task canon event. The Cosim score is cosin similarity between matched $<$thinking$>$-$<$acting$>$ represent trustworthiness of their response.}
\label{tab:model_comparison}
\end{table*}

\subsection{Cross-Character Generalization}
\begin{figure}[ht]
    \centering
    \begin{tikzpicture}
    \begin{axis}[
    ybar=0pt,
    bar width=10pt,
    width=8cm,
    height=9cm,
    ymin=0, ymax=0.8,
    ylabel={Accuracy},
    symbolic x coords={4o-mini, gemini2.5-flash-think,   gemini2-flash, r1, sonnet3.5, sonnet3.7, v3},
    xtick=data,
    x tick label style={rotate=45, anchor=east},
    enlarge x limits=0.15,
    title={Dilemma\_Cross and Canon\_Cross },
    legend style={at={(0.5,-0.2)}, anchor=north, legend columns=-1},
    nodes near coords,
    nodes near coords align={vertical},
    nodes near coords style={
        /pgf/number format/.cd,
        fixed,
        fixed zerofill,
        precision=2
    }
]

\addplot+[fill=blue!30] coordinates {
    (4o-mini,0.6342592593)
    (gemini2.5-flash-think,0.4393518519)
    (gemini2-flash,0.5527777778)
    (r1,0.6240740741)
    (sonnet3.5,0.6888888889)
    (sonnet3.7,0.6)
    (v3,0.5578703704)
};

\addplot+[fill=red!30] coordinates {
    (4o-mini,0.5850668648)
    (gemini2.5-flash-think,0.4231054978)
    (gemini2-flash,0.4513372957)
    (r1,0.6192421991)
    (sonnet3.5,0.6500742942)
    (sonnet3.7,0.5906389302)
    (v3,0.5921248143)
};
\legend{Dilemma Cross, Canon Cross}

\end{axis}
\end{tikzpicture}
    \caption{Accuracy result from cross characters evaluation on task canon and task dilemma.}
    \label{fig:cross}
\end{figure}
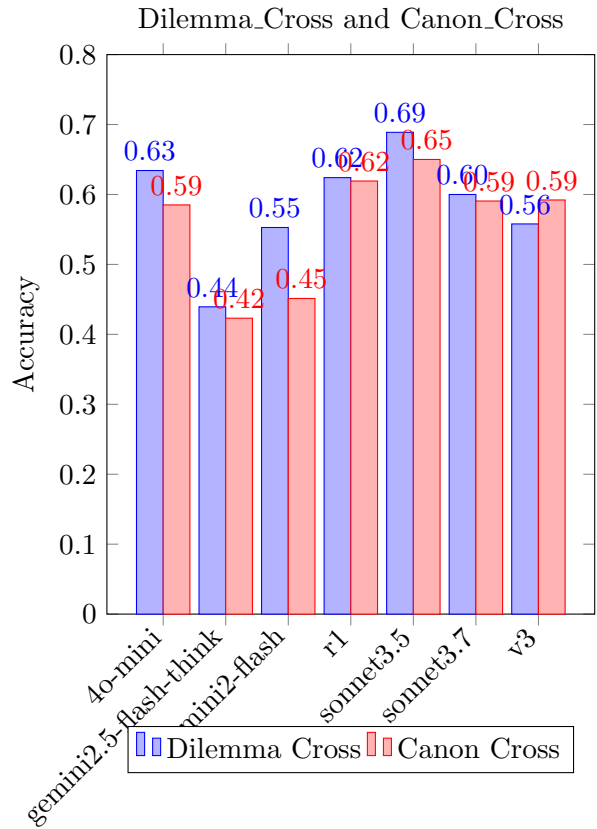

(Figure~\ref{fig:cross}) shows cross-character transfer. Here, accuracy is substantially lower than within-character evaluations, reflecting the difficulty of distinguishing between overlapping but divergent timelines. Sonnet3.5 again achieves the strongest performance across both Dilemma Cross (0.69) and Canon Cross (0.65), highlighting its robustness to shifts in character context. Conversely, gemini2.5-flash-think shows the steepest drop, indicating vulnerability to timeline conflation. This reinforces that multiversal coherence is not trivially solved by scale or reasoning, but requires models to anchor their decisions in fine-grained contextual cues.

\subsection{Thinking vs. Acting}

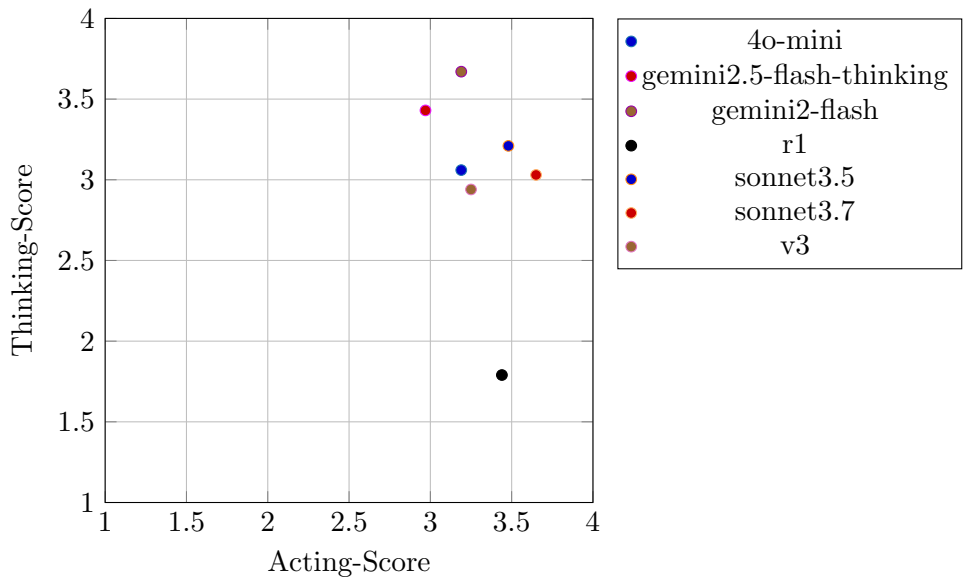
\begin{figure*}[ht]
\centering
\begin{tikzpicture}
\begin{axis}[
    xlabel={Acting-Score},
    ylabel={Thinking-Score},
    width=8cm, height=8cm,
    xmin=1, xmax=4,
    ymin=1, ymax=4,
    grid=both,
    legend style={at={(1.05,1)},anchor=north west}
]
\addplot+[only marks,mark=*,color=model4o] coordinates {(3.19,3.06)}; \addlegendentry{4o-mini}
\addplot+[only marks,mark=*,color=modelgem25] coordinates {(2.97,3.43)}; \addlegendentry{gemini2.5-flash-thinking}
\addplot+[only marks,mark=*,color=modelgem2] coordinates {(3.19,3.67)}; \addlegendentry{gemini2-flash}
\addplot+[only marks,mark=*,color=black] coordinates {(3.44,1.79)}; \addlegendentry{r1}
\addplot+[only marks,mark=*,color=modelson35] coordinates {(3.48,3.21)}; \addlegendentry{sonnet3.5}
\addplot+[only marks,mark=*,color=modelson37] coordinates {(3.65,3.03)}; \addlegendentry{sonnet3.7}
\addplot+[only marks,mark=*,color=modelv3] coordinates {(3.25,2.94)}; \addlegendentry{v3}
\end{axis}
\end{tikzpicture}
\caption{scores for acting-thinking.}
\label{fig:think-act}
\end{figure*}

(Figure~\ref{fig:think-act}) plots the score that is computed in Section~\ref{Reasoning analysis framework}. Clusters in the upper-right quadrant, showing moderate-to-high fidelity in both dimensions. However, different models emphasize different aspects: gemini2 exhibits stronger reasoning (“thinking” score 3.67) but weaker action alignment, while sonnet3.7 scores highest in acting (3.65) yet shows only moderate internal reasoning (3.03). This divergence illustrates the reasoning–acting gap: models may articulate internally consistent deliberations without translating them into persona-faithful actions, or vice versa.

The Think–Act Matching metric provides further nuance. For example, r1 demonstrates the largest increase in reasoning–action alignment under CoT, despite reduced canonical accuracy. This suggests that reasoning traces can improve trustworthiness of role-played responses, even when they do not strictly improve correctness.

\subsection{Discussion}

Taken together, these findings highlight several broader insights:

Reasoning CoT helps, but not uniformly: CoT improves coherence for weaker models but can harm stronger ones by over-generating or straying from canon. This echoes \citep{reasoningdoesnecessarilyimprove} prior findings on CoT’s mixed effects in knowledge-intensive tasks.

Multiversal consistency is especially hard: Even high-performing models confuse character variants, underscoring the value of our benchmark in testing fine-grained temporal and narrative distinctions.

Acting vs Thinking: A model can stay “in character” in surface-level action without justifying its decisions, or it can reason deeply but act inconsistently. Bridging this gap is key for trustworthy role-play agents.

Overall, our benchmark reveals that current LLMs fall short of fully capturing version-specific character portrayals, especially under timeline shifts and moral dilemmas. These results suggest future work should explore integrated reasoning–persona modeling, potentially combining structured world knowledge with dynamic narrative alignment.

\section*{Acknowledgments}
This research was supported by the Faculty of Engineering, Thammasat School of Engineering, Thammasat University also thanks to PreceptorAI that provides API for generate additional training data.

\bibliography{custom}

\begin{thebibliography}{33}
\providecommand{\natexlab}[1]{#1}

\bibitem[{tho()}]{thor_odinson_pdb}

\newblock Personality type - mcu: The heroes.
\newblock \url{https://www.personality-database.com/profile/154/thor-odinson-mcu-the-heroes-mbti-personality-type}.
\newblock Accessed: 2024-08-16.

\bibitem[{Afzoon et~al.(2024)Afzoon, Naseem, Beheshti, and Jamali}]{PersoBench}
Saleh Afzoon, Usman Naseem, Amin Beheshti, and Zahra Jamali. 2024.
\newblock \href {https://arxiv.org/abs/2410.03198} {Persobench: Benchmarking personalized response generation in large language models}.
\newblock \emph{Preprint}, arXiv:2410.03198.

\bibitem[{Ahn et~al.(2024)Ahn, Lee, Lim, Kim, Yun, Lee, and Kim}]{ahn-etal-2024-timechara}
Jaewoo Ahn, Taehyun Lee, Junyoung Lim, Jin-Hwa Kim, Sangdoo Yun, Hwaran Lee, and Gunhee Kim. 2024.
\newblock \href {https://doi.org/10.18653/v1/2024.findings-acl.197} {{T}ime{C}hara: Evaluating point-in-time character hallucination of role-playing large language models}.
\newblock In \emph{Findings of the Association for Computational Linguistics: ACL 2024}, pages 3291--3325, Bangkok, Thailand. Association for Computational Linguistics.

\bibitem[{Bandura(1999)}]{bandura1999moral}
Albert Bandura. 1999.
\newblock Moral disengagement in the perpetration of inhumanities.
\newblock \emph{Personality and Social Psychology Review}, 3(3):193--209.

\bibitem[{Chen et~al.(2025)Chen, Yao, Zou, Hua, Lyu, Ye, Li, and Wang}]{chen2025designguidelinerpaevaluation}
Chaoran Chen, Bingsheng Yao, Ruishi Zou, Wenyue Hua, Weimin Lyu, Yanfang Ye, Toby Jia-Jun Li, and Dakuo Wang. 2025.
\newblock \href {https://arxiv.org/abs/2502.13012} {Towards a design guideline for rpa evaluation: A survey of large language model-based role-playing agents}.
\newblock \emph{Preprint}, arXiv:2502.13012.

\bibitem[{Chen et~al.(2024)Chen, Chen, Yan, Xu, Xing, Shen, Quan, Li, Zhang, and Huang}]{socialbench}
Hongzhan Chen, Hehong Chen, Ming Yan, Wenshen Xu, Gao Xing, Weizhou Shen, Xiaojun Quan, Chenliang Li, Ji~Zhang, and Fei Huang. 2024.
\newblock \href {https://doi.org/10.18653/v1/2024.findings-acl.125} {{S}ocial{B}ench: Sociality evaluation of role-playing conversational agents}.
\newblock In \emph{Findings of the Association for Computational Linguistics: ACL 2024}, pages 2108--2126, Bangkok, Thailand. Association for Computational Linguistics.

\bibitem[{Chen et~al.(2023)Chen, Wang, Jiang, Cai, Li, Chen, Wang, and Li}]{HPD_dataset}
Nuo Chen, Yan Wang, Haiyun Jiang, Deng Cai, Yuhan Li, Ziyang Chen, Longyue Wang, and Jia Li. 2023.
\newblock \href {https://doi.org/10.18653/v1/2023.findings-emnlp.570} {Large language models meet harry potter: A dataset for aligning dialogue agents with characters}.
\newblock In \emph{Findings of the Association for Computational Linguistics: EMNLP 2023}, pages 8506--8520, Singapore. Association for Computational Linguistics.

\bibitem[{{DC Database contributors}(2025)}]{dcau_fandom}
{DC Database contributors}. 2025.
\newblock \href {https://dc.fandom.com/wiki/DCAU} {Dcau}.
\newblock Accessed: 2024-08-19.

\bibitem[{{DC Extended Universe Wiki contributors}(2025)}]{dceu_wiki}
{DC Extended Universe Wiki contributors}. 2025.
\newblock \href {https://dcextendeduniverse.fandom.com} {Dc extended universe wiki}.
\newblock [Online; accessed 19-May-2025].

\bibitem[{deepmind(2025)}]{google2025gemini2.5}
deepmind. 2025.
\newblock Gemini 2.5: Our newest gemini model with thinking.
\newblock \url{https://blog.google/technology/google-deepmind/gemini-model-thinking-updates-march-2025/}.
\newblock Accessed: 2025-05-18.

\bibitem[{DeepSeek-AI et~al.(2025)DeepSeek-AI, Guo, Yang, Zhang, Song, Zhang, Xu, Zhu, Ma, Wang, Bi, Zhang, Yu, Wu, Wu, Gou, Shao, Li, Gao, Liu, Xue, Wang, Wu, Feng, Lu, Zhao, Deng, Zhang, Ruan, Dai, Chen, Ji, Li, Lin, Dai, Luo, Hao, Chen, Li, Zhang, Bao, Xu, Wang, Ding, Xin, Gao, Qu, Li, Guo, Li, Wang, Chen, Yuan, Qiu, Li, Cai, Ni, Liang, Chen, Dong, Hu, Gao, Guan, Huang, Yu, Wang, Zhang, Zhao, Wang, Zhang, Xu, Xia, Zhang, Zhang, Tang, Li, Wang, Li, Tian, Huang, Zhang, Wang, Chen, Du, Ge, Zhang, Pan, Wang, Chen, Jin, Chen, Lu, Zhou, Chen, Ye, Wang, Yu, Zhou, Pan, Li, Zhou, Wu, Ye, Yun, Pei, Sun, Wang, Zeng, Zhao, Liu, Liang, Gao, Yu, Zhang, Xiao, An, Liu, Wang, Chen, Nie, Cheng, Liu, Xie, Liu, Yang, Li, Su, Lin, Li, Jin, Shen, Chen, Sun, Wang, Song, Zhou, Wang, Shan, Li, Wang, Wei, Zhang, Xu, Li, Zhao, Sun, Wang, Yu, Zhang, Shi, Xiong, He, Piao, Wang, Tan, Ma, Liu, Guo, Ou, Wang, Gong, Zou, He, Xiong, Luo, You, Liu, Zhou, Zhu, Xu, Huang, Li, Zheng, Zhu, Ma, Tang, Zha, Yan, Ren, Ren, Sha, Fu, Xu, Xie, Zhang,
  Hao, Ma, Yan, Wu, Gu, Zhu, Liu, Li, Xie, Song, Pan, Huang, Xu, Zhang, and Zhang}]{r1}
DeepSeek-AI, Daya Guo, Dejian Yang, Haowei Zhang, Junxiao Song, Ruoyu Zhang, Runxin Xu, Qihao Zhu, Shirong Ma, Peiyi Wang, Xiao Bi, Xiaokang Zhang, Xingkai Yu, Yu~Wu, Z.~F. Wu, Zhibin Gou, Zhihong Shao, Zhuoshu Li, Ziyi Gao, and 181 others. 2025.
\newblock \href {https://arxiv.org/abs/arXiv:2501.12948} {Deepseek-r1: Incentivizing reasoning capability in llms via reinforcement learning}.

\bibitem[{Erikson(1968)}]{erikson1968identity}
Erik~H Erikson. 1968.
\newblock \emph{Identity: Youth and Crisis}.
\newblock W. W. Norton \& Company.

\bibitem[{Feng et~al.(2025)Feng, Dou, and Kong}]{reasoningdoesnecessarilyimprove}
Xiachong Feng, Longxu Dou, and Lingpeng Kong. 2025.
\newblock \href {https://arxiv.org/abs/2502.16940} {Reasoning does not necessarily improve role-playing ability}.
\newblock \emph{Preprint}, arXiv:2502.16940.

\bibitem[{Festinger(1957)}]{festinger1957cognitive}
Leon Festinger. 1957.
\newblock \emph{A Theory of Cognitive Dissonance}.
\newblock Stanford University Press.

\bibitem[{Frankl(2006)}]{frankl2006man}
Viktor~E Frankl. 2006.
\newblock \emph{Man's Search for Meaning}.
\newblock Beacon Press.

\bibitem[{Jung(1959)}]{jung1959archetypes}
Carl~Gustav Jung. 1959.
\newblock \emph{The Archetypes and the Collective Unconscious}.
\newblock Princeton University Press.

\bibitem[{Kant(1785)}]{kant1785groundwork}
Immanuel Kant. 1785.
\newblock \emph{Groundwork for the Metaphysics of Morals}.
\newblock Cambridge University Press (translated edition).

\bibitem[{Kohlberg(1971)}]{kohlberg1971stages}
Lawrence Kohlberg. 1971.
\newblock Stages of moral development.
\newblock In \emph{Moral Education}, pages 23--92. Univ. of Toronto Press.

\bibitem[{Litz et~al.(2009)}]{litz2009moral}
Brett~T. Litz and 1 others. 2009.
\newblock Moral injury and moral repair in war veterans: A preliminary model and intervention strategy.
\newblock \emph{Clinical Psychology Review}, 29(8):695--706.

\bibitem[{Lu et~al.(2024)Lu, Yu, Zhou, and Zhou}]{lu-etal-2024-large}
Keming Lu, Bowen Yu, Chang Zhou, and Jingren Zhou. 2024.
\newblock \href {https://doi.org/10.18653/v1/2024.acl-long.423} {Large language models are superpositions of all characters: Attaining arbitrary role-play via self-alignment}.
\newblock In \emph{Proceedings of the 62nd Annual Meeting of the Association for Computational Linguistics (Volume 1: Long Papers)}, pages 7828--7840, Bangkok, Thailand. Association for Computational Linguistics.

\bibitem[{{Marvel Animated Universe Wiki contributors}(2025)}]{marvelanimatedwiki}
{Marvel Animated Universe Wiki contributors}. 2025.
\newblock \href {https://marvelanimated.fandom.com/} {Marvel animated universe wiki}.
\newblock Accessed: 2024-08-19.

\bibitem[{{Marvel Cinematic Universe Wiki contributors}(2025{\natexlab{a}})}]{MCUWiki2025}
{Marvel Cinematic Universe Wiki contributors}. 2025{\natexlab{a}}.
\newblock \href {https://marvelcinematicuniverse.fandom.com/} {Marvel cinematic universe wiki}.
\newblock Accessed: 2024-08-19.

\bibitem[{{Marvel Cinematic Universe Wiki contributors}(2025{\natexlab{b}})}]{MCUWikiThor}
{Marvel Cinematic Universe Wiki contributors}. 2025{\natexlab{b}}.
\newblock \href {https://marvelcinematicuniverse.fandom.com/wiki/Thor} {Thor}.
\newblock Accessed: 2024-08-19.

\bibitem[{{Marvel Database Contributors}(2025)}]{marvel_database}
{Marvel Database Contributors}. 2025.
\newblock \href {https://marvel.fandom.com/} {Marvel database}.
\newblock \url{https://marvel.fandom.com/}.
\newblock Accessed: 2024-08-19.

\bibitem[{Maslow(1943)}]{maslow1943theory}
Abraham~H Maslow. 1943.
\newblock A theory of human motivation.
\newblock \emph{Psychological Review}, 50(4):370--396.

\bibitem[{Song et~al.(2020)Song, Tan, Qin, Lu, and Liu}]{all-mpnet-base-v2}
Kaitao Song, Xu~Tan, Tao Qin, Jianfeng Lu, and Tie-Yan Liu. 2020.
\newblock Mpnet: masked and permuted pre-training for language understanding.
\newblock In \emph{Proceedings of the 34th International Conference on Neural Information Processing Systems}, NIPS '20, Red Hook, NY, USA. Curran Associates Inc.

\bibitem[{Stammbach et~al.(2022)Stammbach, Antoniak, and Ash}]{stammbach-etal-2022-heroes}
Dominik Stammbach, Maria Antoniak, and Elliott Ash. 2022.
\newblock \href {https://doi.org/10.18653/v1/2022.wnu-1.6} {Heroes, villains, and victims, and {GPT}-3: Automated extraction of character roles without training data}.
\newblock In \emph{Proceedings of the 4th Workshop of Narrative Understanding (WNU2022)}, pages 47--56, Seattle, United States. Association for Computational Linguistics.

\bibitem[{Tu et~al.(2024)Tu, Fan, Tian, Shen, Shang, Gao, and Yan}]{charactereval}
Quan Tu, Shilong Fan, Zihang Tian, Tianhao Shen, Shuo Shang, Xin Gao, and Rui Yan. 2024.
\newblock \href {https://doi.org/10.18653/v1/2024.acl-long.638} {{C}haracter{E}val: A {C}hinese benchmark for role-playing conversational agent evaluation}.
\newblock In \emph{Proceedings of the 62nd Annual Meeting of the Association for Computational Linguistics (Volume 1: Long Papers)}, pages 11836--11850, Bangkok, Thailand. Association for Computational Linguistics.

\bibitem[{Wang et~al.(2024{\natexlab{a}})Wang, Peng, Que, Liu, Zhou, Wu, Guo, Gan, Ni, Yang, Zhang, Zhang, Ouyang, Xu, Huang, Fu, and Peng}]{wang-etal-2024-rolellm}
Noah Wang, Z.y. Peng, Haoran Que, Jiaheng Liu, Wangchunshu Zhou, Yuhan Wu, Hongcheng Guo, Ruitong Gan, Zehao Ni, Jian Yang, Man Zhang, Zhaoxiang Zhang, Wanli Ouyang, Ke~Xu, Wenhao Huang, Jie Fu, and Junran Peng. 2024{\natexlab{a}}.
\newblock \href {https://doi.org/10.18653/v1/2024.findings-acl.878} {{R}ole{LLM}: Benchmarking, eliciting, and enhancing role-playing abilities of large language models}.
\newblock In \emph{Findings of the Association for Computational Linguistics: ACL 2024}, pages 14743--14777, Bangkok, Thailand. Association for Computational Linguistics.

\bibitem[{Wang et~al.(2024{\natexlab{b}})Wang, Xiao, Huang, Yuan, Xu, Guo, Tu, Fei, Leng, Wang, Chen, Li, and Xiao}]{wang-etal-2024-incharacter}
Xintao Wang, Yunze Xiao, Jen-tse Huang, Siyu Yuan, Rui Xu, Haoran Guo, Quan Tu, Yaying Fei, Ziang Leng, Wei Wang, Jiangjie Chen, Cheng Li, and Yanghua Xiao. 2024{\natexlab{b}}.
\newblock \href {https://doi.org/10.18653/v1/2024.acl-long.102} {{I}n{C}haracter: Evaluating personality fidelity in role-playing agents through psychological interviews}.
\newblock In \emph{Proceedings of the 62nd Annual Meeting of the Association for Computational Linguistics (Volume 1: Long Papers)}, pages 1840--1873, Bangkok, Thailand. Association for Computational Linguistics.

\bibitem[{Wei et~al.(2022)Wei, Wang, Schuurmans, Bosma, ichter, Xia, Chi, Le, and Zhou}]{COT}
Jason Wei, Xuezhi Wang, Dale Schuurmans, Maarten Bosma, brian ichter, Fei Xia, Ed~Chi, Quoc~V Le, and Denny Zhou. 2022.
\newblock \href {https://proceedings.neurips.cc/paper_files/paper/2022/file/9d5609613524ecf4f15af0f7b31abca4-Paper-Conference.pdf} {Chain-of-thought prompting elicits reasoning in large language models}.
\newblock In \emph{Advances in Neural Information Processing Systems}, volume~35, pages 24824--24837. Curran Associates, Inc.

\bibitem[{Xi et~al.(2023)Xi, Chen, Guo, He, Ding, Hong, Zhang, Wang, Jin, Zhou, Zheng, Fan, Wang, Xiong, Zhou, Wang, Jiang, Zou, Liu, Yin, Dou, Weng, Cheng, Zhang, Qin, Zheng, Qiu, Huang, and Gui}]{xi2023risepotentiallargelanguage}
Zhiheng Xi, Wenxiang Chen, Xin Guo, Wei He, Yiwen Ding, Boyang Hong, Ming Zhang, Junzhe Wang, Senjie Jin, Enyu Zhou, Rui Zheng, Xiaoran Fan, Xiao Wang, Limao Xiong, Yuhao Zhou, Weiran Wang, Changhao Jiang, Yicheng Zou, Xiangyang Liu, and 10 others. 2023.
\newblock \href {https://arxiv.org/abs/2309.07864} {The rise and potential of large language model based agents: A survey}.
\newblock \emph{Preprint}, arXiv:2309.07864.

\bibitem[{Zhang et~al.(2018)Zhang, Dinan, Urbanek, Szlam, Kiela, and Weston}]{zhang-etal-2018-personalizing}
Saizheng Zhang, Emily Dinan, Jack Urbanek, Arthur Szlam, Douwe Kiela, and Jason Weston. 2018.
\newblock \href {https://doi.org/10.18653/v1/P18-1205} {Personalizing dialogue agents: {I} have a dog, do you have pets too?}
\newblock In \emph{Proceedings of the 56th Annual Meeting of the Association for Computational Linguistics (Volume 1: Long Papers)}, pages 2204--2213, Melbourne, Australia. Association for Computational Linguistics.

\end{thebibliography}

\clearpage
\appendix
\section*{Appendix}
\addcontentsline{toc}{section}{Appendix}

\section{Our Benchmark}
\label{Our Benchmark}
At the end Our Benchmark consist of three tasks: \textbf{1.Canon Event}(Childhood question for 270, Pre-hero for 270, and hero for 806), \textbf{2.Dilemma Situation}(“Save one vs. Save the greater good” 270 question, “Hero or Villain” 270 , “Duty vs. Desire 270 ” and “Moral Code vs. Outcome 270 ”).

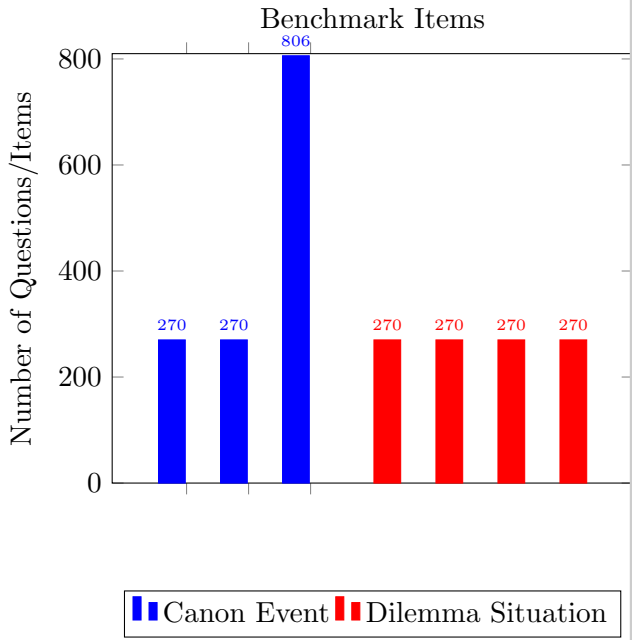
\begin{figure}[ht]
\centering
\begin{tikzpicture}
\begin{axis}[
    title={Benchmark Items},
    ylabel={Number of Questions/Items},
    ybar=1pt, 
    bar width=10pt, 
    ymin=0,
    ymax=810, 
    enlarge x limits=0.2, 
    legend style={
        at={(0.5,-0.25)}, 
        anchor=north,
        legend columns=-1 
    },
    symbolic x coords={
        CE_Childhood, CE_PreHero, CE_Hero, 
        DS_SaveOne, DS_HeroVillain, DS_DutyDesire, DS_MoralOutcome,
        MD_LongContext
    },
    xtick=data, 
    x tick label style={
        rotate=30, 
        anchor=east, 
        font=\footnotesize
    },
    xticklabels={ 
        , , ,
    },
    nodes near coords, 
    nodes near coords align={vertical},
    every node near coord/.append style={font=\tiny, /pgfplots/coordinate style/.append style={yshift=2pt}}
]
\addplot+[style={fill=blue, mark=none}] coordinates {
    (CE_Childhood, 270) 
    (CE_PreHero, 270) 
    (CE_Hero, 806)
};
\addlegendentry{Canon Event}

\addplot+[style={fill=red, mark=none}] coordinates {
    (DS_SaveOne, 270) 
    (DS_HeroVillain, 270) 
    (DS_DutyDesire, 270) 
    (DS_MoralOutcome, 270)
};
\addlegendentry{Dilemma Situation}
\end{axis}
\end{tikzpicture}

\caption{Barplot of our benchmark tasks.The Blue one consist of Childhood,Pre-Hero and Hero question type,the Red one is balanced Dilemma situation.}
\label{fig:benchmark_distribution}
\end{figure}

\section{Model Inference Configs}
\label{sec:Model Inference Configs}

\begin{table}[ht]
    \centering
    \setlength{\tabcolsep}{3pt}  
    \small  
    \begin{tabular}{l c c c}
        \toprule
        Model & Temp & OutToken & CtxLen\\
        \midrule
        GPT-4o-mini & 0.6 & 1024 & 128K\\
        Gemini 2.0-Flash & 0.6 & 1024 & 1M\\
        Gemini-2.5-Flash-Thinking & 0.6 & 1024 & 1M\\
        Sonnet 3.5 & 0.6 & 1024 & 200k\\
        Sonnet 3.7 & 0.6 & 1024 & 200k\\
        DeepSeek r1 & 0.6 & 1024 & 128k\\
        Sonnet 3.7\textbf{(judge)} & 0.1 & 1024 & 200k\\
        \bottomrule
    \end{tabular}
    \caption{Inference Configs for Answer.}
    \label{tab:llms configs answer}
\end{table}

\begin{table}[ht]
    \centering
    \setlength{\tabcolsep}{3pt}  
    \small  
    \begin{tabular}{l c c c}
        \toprule
        Model & Temp & OutToken & CtxLen\\
        \midrule
        GPT-4o-mini\textbf{(tags)} & 0.1 & 1024 & 128K\\
        Gemini 2.0-Flash & 0.9 & 1024 & 1M\\
        Gemini 2.5-Flash-Thinking & 0.9 & 1024 & 1M\\
        Sonnet 3.7 & 0.9 & 1024 & 200k\\
        \bottomrule
    \end{tabular}
    \caption{Inference Configs for Generation.}
    \label{tab:llms configs generation}
\end{table}

\section{Prompts}
\label{sec:Prompts}
\subsection{Dilemma Generation}
\label{Dilemma Generation}
\begin{PromptBox}{Dilemma Generation}
You are creating a complex moral dilemma for \textbf{$<$name$>$} from \textbf{$<$lore$>$}.  

The dilemma type is: \textbf{$<$dilemma\_type$>$}.  

Description of this dilemma type: \textbf{$<$dilemma\_descriptions$>$}  
Create a binary (two-choice) moral dilemma specific to this character's background, abilities, and moral framework.  The dilemma should force the character to make a difficult choice between two valid but conflicting options.  

Format your response as a JSON object with these fields:  
1. "situation" -- Detailed description of the dilemma scenario (3 sentences)  
2. "choice\_A" -- First option (1 sentence)  
3. "choice\_B" -- Second option (1 sentence)  
4. "consequence\_A" -- Consequence of choosing A (1 sentence)  
5. "consequence\_B" -- Consequence of choosing B (1 sentence)  

Make the dilemma deeply personal to this character and challenging based on their specific values and story.  

Make sure to return a valid JSON object without any additional text before or after.
\end{PromptBox}

\subsection{Task prompt}
\label{Task prompt}
\paragraph{Canon Task Prompting:}
\begin{PromptBox}{Canon Task}
    You are playing the role of \textbf{name}, act and think as \textbf{name}, from \textbf{lore}
    $<$question$>$ \textbf{{canonical\_question}}
    \textbf{[option\_A]}, \textbf{[option\_B]},\textbf{[option\_C]}, \textbf{[option\_D]}
    $<$question/$>$
\end{PromptBox}
\paragraph{Dilemma Task Prompting:}
\begin{PromptBox}{Dilemma Task}
    You are playing the role of \textbf{$<$name$>$}, act and think as \textbf{$<$name$>$}, from \textbf{$<$lore$>$}, the situation is \textbf{[dilemma\_scenario]} Choice A: \textbf{[choice\_a]} Consequence A: \textbf{[consequence\_a]} Choice B: \textbf{[choice\_b]} Consequence B: \textbf{[consequence\_b]} $<$question$>$ \textbf{[dilemma\_type\_question]} $<$question/$>$
\end{PromptBox}

\subsection{Reasoning Analysis}
\label{Reasoning Analysis}
\begin{PromptBox}{Judge Scoring}
    "Help me scoring character role-playing, score point between 0-5 , score have two type: thinking(doec thiking response look like reference character) ,acting(does response acting like the character reference), the output must be this format: think\_score,act\_score example 3,2
    Character: \textbf{$<$name$>$} Source: \textbf{$<$source$>$} Attributes: \textbf{$<$attributes$>$} Response to evaluate: \textbf{$<$text\_to\_process$>$}"
    
\end{PromptBox}

\section{Full Results}
\label{sec:Full Results}
Table~\ref{Canon Event Benchmark Results}, Table~\ref{Dilemma Situations Benchmark Results}, Table~\ref{Accuracy results on canon cross}, Table~\ref{Accuracy results on dilemma cross}

\begin{table*}[ht]
\centering
\begin{adjustbox}{max width=\textwidth}
\begin{tabular}{l l r r r r r r r r r r r r}
\toprule
Task & Model & Acc & Prec. & Recall & F1 & Judge\_Act & Judge\_Think & Cosim & Childhood & Pre-Hero & Hero1 & Hero2 & Hero3 \\
\midrule
\textbf{canon} & 4o-mini & 0.626 & 0.646 & 0.626 & 0.632 & 3.191 & 3.064 & 0.458 & 0.581 & 0.641 & 0.663 & 0.593 & 0.650 \\
& gemini2.5-flash-think & 0.663 & 0.682 & 0.663 & 0.665 & 2.972 & 3.429 & 0.450 & 0.700 & 0.667 & 0.644 & 0.689 & 0.613 \\
& gemini2-flash & 0.638 & 0.669 & 0.638 & 0.647 & 3.191 & 3.673 & 0.450 & 0.604 & 0.648 & 0.622 & 0.681 & 0.635 \\
& r1 & 0.652 & 0.667 & 0.652 & 0.656 & 3.443 & 1.789 & 0.296 & 0.622 & 0.733 & 0.622 & 0.656 & 0.628 \\
& sonnet3.5 & 0.704 & 0.714 & 0.704 & 0.707 & 3.475 & 3.211 & 0.433 & 0.674 & 0.733 & 0.700 & 0.704 & 0.711 \\
& sonnet3.7 & 0.637 & 0.671 & 0.637 & 0.645 & 3.652 & 3.025 & 0.378 & 0.630 & 0.656 & 0.622 & 0.656 & 0.624 \\
& v3 & 0.615 & 0.648 & 0.615 & 0.623 & 3.250 & 2.936 & 0.423 & 0.607 & 0.681 & 0.596 & 0.604 & 0.586 \\
\midrule
\textbf{canon\_cot} & 4o-mini & 0.646 & 0.680 & 0.646 & 0.656 & 3.183 & 2.653 & 0.432 & 0.626 & 0.648 & 0.630 & 0.637 & 0.692 \\
& gemini2.5-flash-think & 0.388 & 0.463 & 0.388 & 0.344 & 2.578 & 3.017 & 0.389 & 0.352 & 0.378 & 0.444 & 0.381 & 0.383 \\
& gemini2-flash & 0.637 & 0.698 & 0.637 & 0.652 & 2.661 & 3.443 & 0.408 & 0.615 & 0.663 & 0.637 & 0.641 & 0.632 \\
& r1 & 0.546 & 0.641 & 0.546 & 0.573 & 2.211 & 1.423 & 0.371 & 0.552 & 0.585 & 0.519 & 0.515 & 0.560 \\
& sonnet3.5 & 0.651 & 0.688 & 0.651 & 0.659 & 3.652 & 3.025 & 0.404 & 0.622 & 0.659 & 0.630 & 0.674 & 0.669 \\
& sonnet3.7 & 0.620 & 0.656 & 0.620 & 0.628 & 3.405 & 3.433 & 0.380 & 0.607 & 0.626 & 0.622 & 0.630 & 0.613 \\
& v3 & 0.502 & 0.569 & 0.502 & 0.497 & 3.250 & 2.233 & 0.376 & 0.544 & 0.489 & 0.478 & 0.500 & 0.500 \\
\bottomrule
\end{tabular}
\end{adjustbox}
\caption{Canon Event Benchmark Results}
\label{Canon Event Benchmark Results}
\end{table*}

\begin{table*}[ht]
\centering
\begin{adjustbox}{max width=\textwidth}
\begin{tabular}{l l r r r r r r r r r r r}
\toprule
Task & Model & Acc & Prec. & Recall & F1 & Judge\_Act & Judge\_Think & Cosim & Save\_Love/Good & Hero/Villain & Duty/Desire & Ends/Code \\
\midrule
\textbf{dilemma} & 4o-mini & 0.696 & 0.731 & 0.696 & 0.704 & 3.302 & 3.651 & 0.526 & 0.548 & 0.759 & 0.707 & 0.770 \\
& gemini2.5-flash-think & 0.474 & 0.593 & 0.474 & 0.518 & 2.850 & 3.837 & 0.317 & 0.452 & 0.433 & 0.522 & 0.489 \\
& gemini2-flash & 0.596 & 0.663 & 0.638 & 0.616 & 3.446 & 3.666 & 0.468 & 0.481 & 0.637 & 0.644 & 0.622 \\
& r1 & 0.715 & 0.732 & 0.715 & 0.709 & 3.382 & 2.405 & 0.292 & 0.593 & 0.778 & 0.741 & 0.748\\
& sonnet3.5 & 0.660 & 0.686 & 0.660 & 0.672 & 3.723 & 3.782 & 0.492 & 0.574 & 0.704 & 0.670 & 0.693 \\
& sonnet3.7 & 0.571 & 0.684 & 0.571 & 0.618 & 3.892 & 4.070 & 0.400 & 0.507 & 0.511 & 0.648 & 0.619 \\
& v3 & 0.563 & 0.602 & 0.615 & 0.623 & 3.638 & 3.744 & 0.420 & 0.552 & 0.648 & 0.559 & 0.493 \\
\midrule
\textbf{dilemma\_cc} & 4o-mini & 0.669 & 0.708 & 0.669 & 0.684 & - & - & - & 0.481 & 0.778 & 0.700 & 0.719 \\
& gemini2.5-flash-think & 0.459 & 0.575 & 0.459 & 0.505 & - & - & - & 0.470 & 0.430 & 0.515 & 0.422 \\
& gemini2-flash & 0.569 & 0.664 & 0.569 & 0.610 & - & - & - & 0.485 & 0.611 & 0.485 & 0.536 \\
& r1 & 0.707 & 0.720 & 0.707 & 0.703 & - & - & - & 0.619 & 0.763 & 0.730 & 0.719 \\
& sonnet3.5 & 0.684 & 0.720 & 0.684 & 0.694 & - & - & - & 0.604 & 0.685 & 0.711 & 0.737 \\
& sonnet3.7 & - & - & - & - & - & - & - & - & - & - & - \\
& v3 & 0.563 & 0.608 & 0.563 & 0.578 & - & - & - & 0.463 & 0.619 & 0.567 & 0.604 \\
\bottomrule
\end{tabular}
\end{adjustbox}
\caption{Dilemma Situations Benchmark Results}
\label{Dilemma Situations Benchmark Results}
\end{table*}

\begin{table}[ht]
\centering
\label{tab:canon_cross_results}
\begin{tabular}{l c}
\toprule
\textbf{Model} & \textbf{Accuracy} \\
\midrule
4o-mini                 & 0.585 \\
gemini2.5-flash-think   & 0.423 \\
gemini2-flash           & 0.451 \\
r1                      & 0.619 \\
sonnet3.5               & 0.650 \\
sonnet3.7               & 0.591 \\
v3                      & 0.592 \\
\bottomrule
\end{tabular}
\caption{Accuracy results on canon cross characters evaluation.}
\label{Accuracy results on canon cross}
\end{table}

\begin{table}[ht]
\centering
\label{tab:dilemma_cross_results}
\begin{tabular}{l c}
\toprule
\textbf{Model} & \textbf{Accuracy} \\
\midrule
4o-mini                 & 0.634 \\
gemini2.5-flash-think   & 0.439 \\
gemini2-flash           & 0.553 \\
r1                      & 0.624 \\
sonnet3.5               & 0.689 \\
sonnet3.7               & 0.600 \\
v3                      & 0.558 \\
\bottomrule
\end{tabular}
\caption{Accuracy results on dilemma cross characters evaluation.}
\label{Accuracy results on dilemma cross}
\end{table}

\section{Hero Character Profiles}
\label{sec:Hero Character Profiles}

\subsection{Hero Selection}
\label{sec:Hero Selection}
In this study, we selected superhero characters from various universes within the Marvel and DC franchises, including the Marvel Cinematic Universe (MCU)~\cite{MCUWiki2025}, the Marvel Animated Universe~\cite{marvelanimatedwiki}, the Marvel Database~\cite{marvel_database}, the DC Extended Universe (DCEU)~\cite{dceu_wiki}, and the DC Animated Universe (DCAU)~\cite{dcau_fandom}. The selection process involved filtering for heroes with well-documented backstories who consistently demonstrate pure-hearted intentions and are not categorized as anti-heroes. These criteria were applied to ensure coherence with the scope and objectives of the present research.

\subsection{Hero Attributes}
\label{sec:Hero Attributes}

\begin{figure}[ht]
    \centering
    \includegraphics[width=1\linewidth]{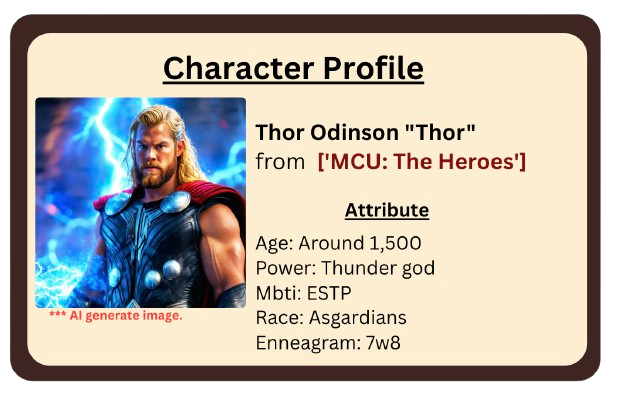}
    \caption{Example of a character profile for Thor Odinson (Thor) from the MCU: The Heroes universe, illustrating the attribute information associated with the character within this universe.}
    \label{fig:thor_character_profile}
\end{figure}

\begin{figure}[ht]
    \centering
    \includegraphics[width=1\linewidth]{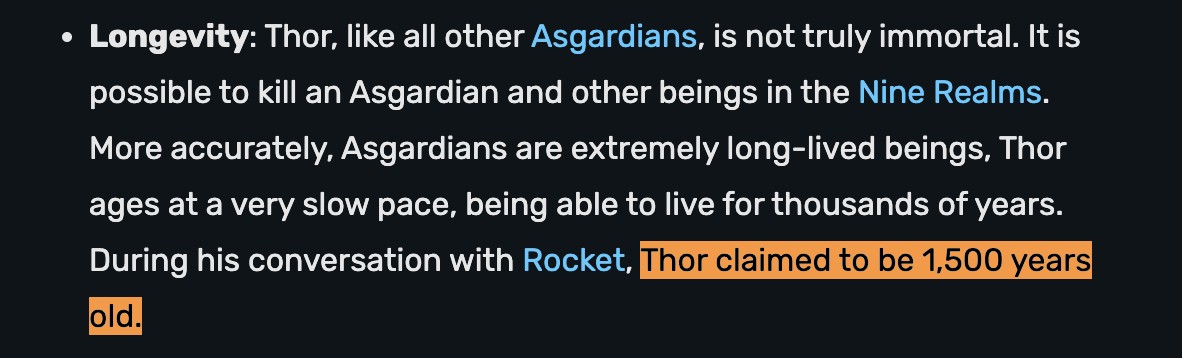}
    \caption{Character age information of Thor Odinson (Thor) from the MCU: The Heroes universe. The age is explicitly referenced in Avengers: Infinity War, where Thor states that he is approximately 1,500 years old.~\cite{MCUWikiThor}}
    \label{fig:thor_age}
\end{figure}

\begin{figure}[ht]
    \centering
    \includegraphics[width=0.7\linewidth]{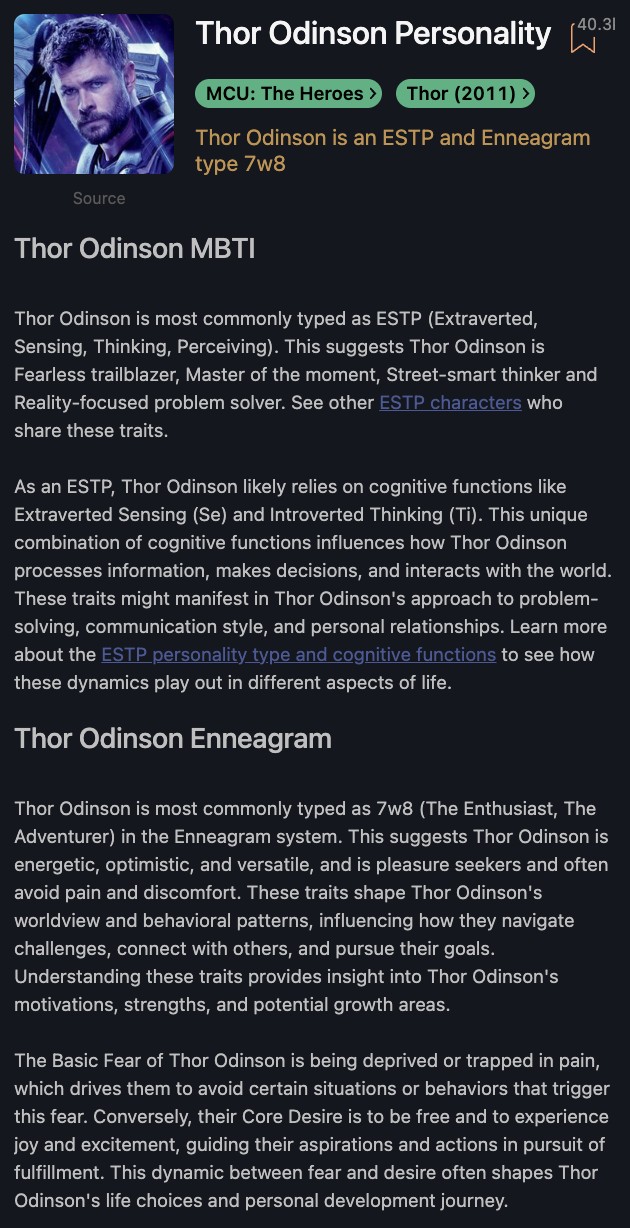}
    \caption{Character profile webpage of Thor Odin- son (Thor) from the MCU: The Heroes universe from the Personality Database (PDB). This figure shows the online personality profile of Thor Odinson, as portrayed in the Marvel Cinematic Universe (MCU), according to user-contributed data on the Personality Database website. The profile includes MBTI, Enneagram, and other personality traits derived from fan-based assessments.~\cite{thor_odinson_pdb}}
    \label{fig:thor_pdb}
\end{figure}

During the development of character attributes for the hero evaluation section, we identified five core characteristics: (1) age - the age of the character within the universe up to the latest film or series episode; (2) Power – the hero's abilities; (3) MBTI – the character’s Myers-Briggs Type Indicator; (4) Race – the character’s species or lineage; and (5) Enneagram – the character’s Enneagram type.
General character information, including Age, Power, and Race, was sourced from official and community-curated Wiki fandom pages of each universe. Psychological traits such as MBTI and enneagram types were referenced from PDB: The Personality Database.
For example, the character profile of Thor Odinson from the Marvel Cinematic Universe (MCU), shown in Figure~\ref{fig:thor_character_profile}, lists his age as 'around 1,500 years', a detail revealed in Avengers: Infinity War and documented on the MCU Wiki (Figure~\ref{fig:thor_age}). Similarly, Thor's psychological traits, MBTI and Enneagram, were derived from user-contributed assessments in the Personality Database (Figure~\ref{fig:thor_pdb}).


\subsection{Question Constructions}
\label{Question Constructions}
\begin{figure}[ht]
    \centering
    \includegraphics[width=1.0\linewidth]{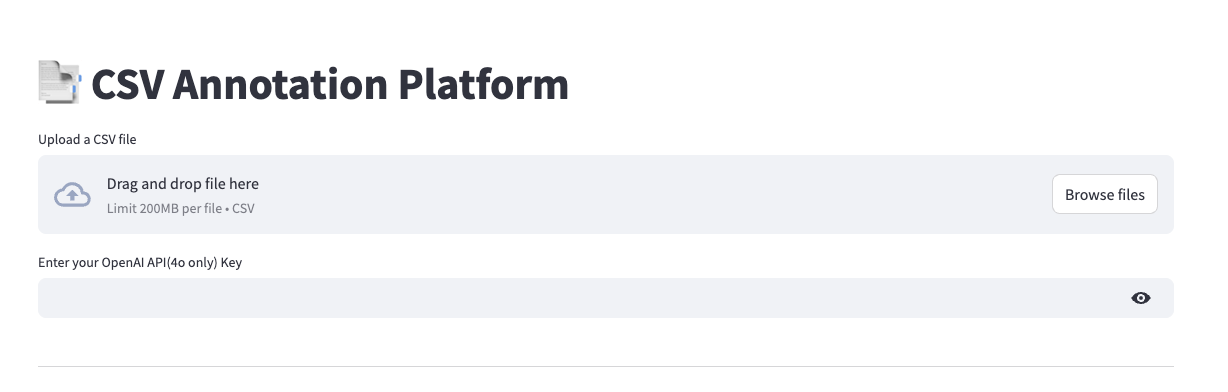}
    \includegraphics[width=1.0\linewidth]{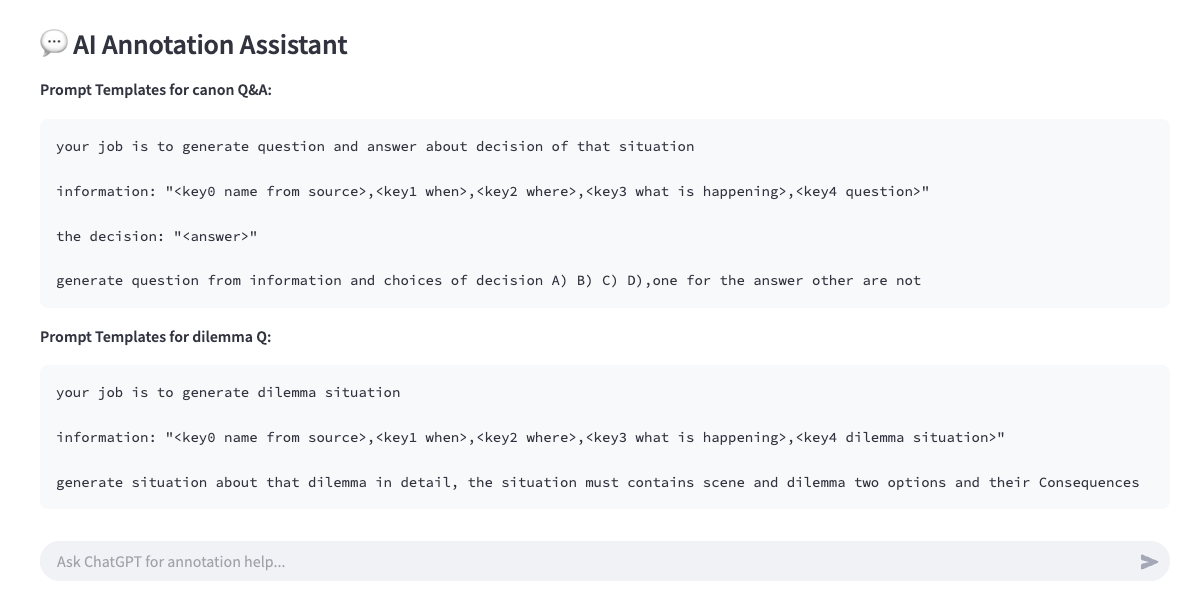}
    \caption{Platform to help experts construct question and answer to canon event in multiple choices format and construct question in dilemma situation}
    \label{fig:platform_construct}
\end{figure}

To construct a comprehensive benchmark dataset for evaluating.This process involved capturing key aspects of character knowledge and decision-making capabilities in LLM. We developed a custom annotation platform and defined specific structures for leveraging both AI assistance(GPT-4o-mini) and human experts oversight. 
An annotation platform was developed using \textbf{Streamlit} 
framework as Figure\ref{fig:platform_construct}. This platform provided a user-friendly
interface. It presented expert annotators with structured templates
designed to capture information systematically for different benchmarking scenarios. Based on the structured information by the expert Crucially, while GPT-4o-mini assisted in generation, the \textbf{final validation, refinement, and approval of all data points depends 
on our expert annotators.}

\subsubsection{Generation Structure}
The annotation process centered around two primary structures.

\paragraph{Canon Event Structure:} This structure format focused on evaluating the model’s ability to recall and accurately represent established facts and events from a character’s history in their lore, by creating multiple-choice Q\&A. For each canon event entry, expert annotators were required to complete the following fields within the platform:

\begin{itemize}
  \item \textbf{Character Name and Lore:} The name of the hero character and their universe name.
  \item \textbf{Time (When):} The specific time context in which the event occurred.
  \item \textbf{Location (Where):} The setting or place where the event took place.
  \item \textbf{Description (What happened):} A factual account of a significant event from the character’s storyline (canon).
  \item \textbf{Question:} A question pertinent to the key event, designed to test the model’s knowledge. This question must be initially drafted by experts.
  \item \textbf{One True Answer:} One canonically correct answer by experts and the other answers are generated by GPT-4o-mini.
\end{itemize}

\paragraph{Dilemma Structure:} This structure aimed to assess the model’s capacity for nuanced role-playing, specifically in navigating complex situations that require decision-making consistent with the character’s established personality, morals, and values. Experts filled out the following fields for each dilemma scenario:

\begin{itemize}
  \item \textbf{Character Name and Lore:} The name of the hero character and their universe (which movie or comic).
  \item \textbf{Situation – Time (When):} The temporal context for the dilemma.
  \item \textbf{Situation – Location (Where):} The setting where the dilemma unfolds.
  \item \textbf{Situation – Context (What is happening):} Background information setting the stage and explaining the circumstances leading to the dilemma situation.
  \item \textbf{Dilemma Type:} A type of dilemma such as ``Save vs. Sacrifice'', ``Hero or Villain'', ``Duty vs. Desire''.
\end{itemize}

\end{document}